\documentclass[journal]{IEEEtran}
\usepackage{amsmath}
\usepackage{amssymb}
\usepackage{graphicx}
\usepackage[table]{xcolor}
\usepackage{xcolor}
\usepackage{todonotes}
\usepackage{caption}
\usepackage{subcaption}
\usepackage[export]{adjustbox}
\usepackage{cite}
\usepackage[textsize=tiny,draft]{changes}
\ifCLASSINFOpdf
\else
\fi
\begin{document}
%
\title{Dynamic  Control of Soft Robotic Arm}
%
%
%

\author{Milad~Azizkhani$^{1}$,
         Isuru~S. Godage$^{2}$,
        and Yue~Chen$^{3}$
\thanks{This research was partially funded by National Science Foundation grant IIS-1718755. Corresponding author: Yue Chen}
\thanks{$^{1}$Milad Azizkhani is with the Department of Mechanical Engineering, University of Arkansas, Fayetteville, AR, 72701. {\tt\small ma118@uark.edu}}
\thanks{$^{2}$Isuru~S.~Godage is with the College of Computing and Digital media, DePaul University,
        Chicago, IL, 60604
        {\tt\small igodage@depaul.edu}}
\thanks{$^{3}$Yue Chen is with the Department of Biomedical Engineering, Georgia Institute of Technology/Emory University, Atlanta, GA, 30332.
        {\tt\small yue.chen@bme.gatech.edu}}%
\thanks{This work has been submitted to the IEEE for possible publication. Copyright may be transferred without notice, after which this version may no longer be
accessible.}
}

\maketitle

\begin{abstract}
In this article, the control problem of one section pneumatically actuated soft robotic arm is investigated in detail. To date, extensive prior work has been done in soft robotics kinematics and dynamics modeling. Proper controller designs can complement the modeling part since they are able to compensate other effects that have not been considered in the modeling, such as the model uncertainties, system parameter identification error, hysteresis, etc. In this paper, we explored different control approaches (kinematic control, PD+feedback linearization, passivity control, adaptive passivity control) and summarized  the advantages and disadvantages of each controller. 
We further investigated the robot control problem in the practical scenarios when the sensor noise exists, actuator velocity measurement is not available, and the hysteresis effect is non-neglectable. Our simulation results indicated that the adaptive passivity control with sigma modification terms, along with a high-gain observer presents a better performance in comparison with other approaches. Although this paper mainly presented the simulation results of various controllers, the work will pave the way for practical implementation of soft robot control. 

\end{abstract}


%
\IEEEpeerreviewmaketitle

\section{Introduction}

%
%
%
%
\IEEEPARstart{S}{oft} robots for their continuum structure and  intrinsic safety in human-robot interaction have attracted increasing attention in recent years \cite{laschi2016soft}. These robots are mainly made of soft and elastic materials such as silicone-rubber \cite{rus2015design}, which makes them compliant to unstructured environments\cite{della2020model}. The actuation approaches for soft robots can be different, and these robots can be actuated using shape memory alloy \cite{jiang2020variable}, magnetic \cite{bira2020review}, pneumatic-driven \cite{chen2021modal}, fluid-driven \cite{caldwell1995control},  tendon-drive \cite{li2021development}, and many other approaches
\cite{rus2015design}.
%
Out of numerous actuation technologies, pneumatics remains the most popular due to their ease of adoption, high power-to-weight ratio, low cost, customizability, high bandwidth, and rapid producibility.
Soft robotic arms are typically constructed by serially connecting two or more soft bending units (i.e., sections). Each section is powered by multiple variable-length soft actuators. Differential length of these actuators -- due to the difference in input pressures -- results in a circular arc-shape deformation, which is usually parameterized by arc parameters.
These soft robots have been used in many emerging areas such as rehabilitation \cite{POLYGERINOS2015135}, surgery \cite{li2021development}, harvesting \cite{gunderman2021tendon}, etc.

Despite their advantages and potentials, soft robots' control remains a challenging task. This challenge consists of several aspects. First of all, the model implemented into the controller should be derived in a way to take into account the continuous nature, coupled-nonlinear  dynamics, and external forces of the system.
In addition, there are other factors such as hysteresis, nonlinear/unmodelled dynamics, parametric uncertainty that made it even more challenging. 
%
%
%

Cosserat rod theory \cite{till2019real} and Finite elements methods (FEM)\cite{polygerinos2015modeling} can describe the system dynamics with great accuracy. 
%
The real-time implementation thereof challenging has only been possible for simpler robots with a few degrees of freedom (DoF) due to coupled, computationally expensive equations governing the dynamics.
%
Thus, their utility in real-time dynamic controller design is challenging.
In addition, the format of their dynamic equations makes the designing of nonlinear robust adaptive controllers difficult \cite{alqumsan2019robust}. 
Alqumsan \textit{et al.} \cite{alqumsan2019robust} have addressed this problem by proposing a sliding mode control based on the $\alpha$-method which is an implicit numerical method. 
%
Further, controlling the robot with Cosserat rod model or FEM  is computationally expensive and thus was only evaluated numerically or performed in quasi-static applications. 
%

In contrast, piece-wise constant curvature (PCC) modeling approach employs simplifying assumptions \cite{falkenhahn2015dynamic} -- which are valid for the vast majority of manipulation applications -- to reduce the computational burden. 
It should be noted that the assumptions should be always valid, otherwise, the model would not be accurate enough to describe the system   dynamics\cite{renda2014dynamic}. Thus, lumped approaches circumvent the modeling complexities inherent in soft robotic arms to achieve efficient results \cite{khalil2007dynamic}. However, the need for many discrete units to model smooth bending results in inefficient models, unsuitable for real-time control implementation. 
%
Based on the modal kinematics -- that combines the accuracy, efficiency, and stability -- the spatial dynamic model reported in \cite{godage2016dynamics} demonstrated real-time simulations for a 3-section soft robotic arm. In addition, the center-of-gravity-based model \cite{godage2019center} further improved the numerical efficiency and resulted in sub-real-time spatial dynamic models rendering such model suitable for real-time closed-loop controller design. Despite the recent advances in dynamic modeling, research progress in controllers still lags.

%
The work reported in \cite{dehghani2011modeling} is one of the early attempts to derive controllers for a planar soft manipulator. A controller for a tendon-driven manipulator presented in \cite{li2018design} reported tracking of simple spatial trajectories. Given the slower tracking trajectories, absence of nonlinear effects (friction, hysteresis), and lack of details regarding the dynamic stability thereof available, limits the viability of the model to control faster soft manipulators. In other studies\cite{FalkenControl,della2020model}, feedback linearization with proportional derivative (PD) controller has have been implemented based on the accurate modeling of the robot. However, the effect of uncertainty and hysteresis is not addressed in the control. The dynamic control design for a soft pneumatic actuator in the presence of hysteresis effect is addressed in \cite{godage2018dynamic}. In addition, the model-based control of continuum soft robots in the presence of uncertainty in the system are addressed in \cite{kapadia2010model,Trumic}.

In this paper, we aim to present a new control approach that not only takes into account the known dominant dynamics of the robot but also maintain the superior trajectory tracking in the presence of parametric uncertainty, hysteresis, sensor noise, unavailability of actuators' velocity measurements, and unmodeled dynamics. The rest of the paper is organized as follows. In section \ref{sec:model}, the kinematic and dynamic models of the robot were briefly introduced. In section \ref{sec:controlcomparison}, different control strategies were introduced and their performances to track a desired path were presented in section \ref{sec:controlcomparison}. Based on these comparative studies, we discussed the adaptive passivity control with sigma modification to address the limitations during the practical scenarios in section \ref{sec:controlimplementation}. 
Finally,  section \ref{sec:conclusion} presented the conclusion of this paper. 

\section{Soft Robot Modeling: Kinematic and Dynamic}
\label{sec:model}
To implement a model-based control scheme, the kinematic and dynamic model of the soft robotic system should be derived. Note that the kinematics and dynamics model of the soft robot have been developed in our prior work\cite{godage2016dynamics}. In this section, we will give a brief overview of the derivation of these models.
\subsection{Kinematics Modeling}
\begin{figure}[tbh]
    \centering
    \includegraphics[width = 0.38\textwidth]{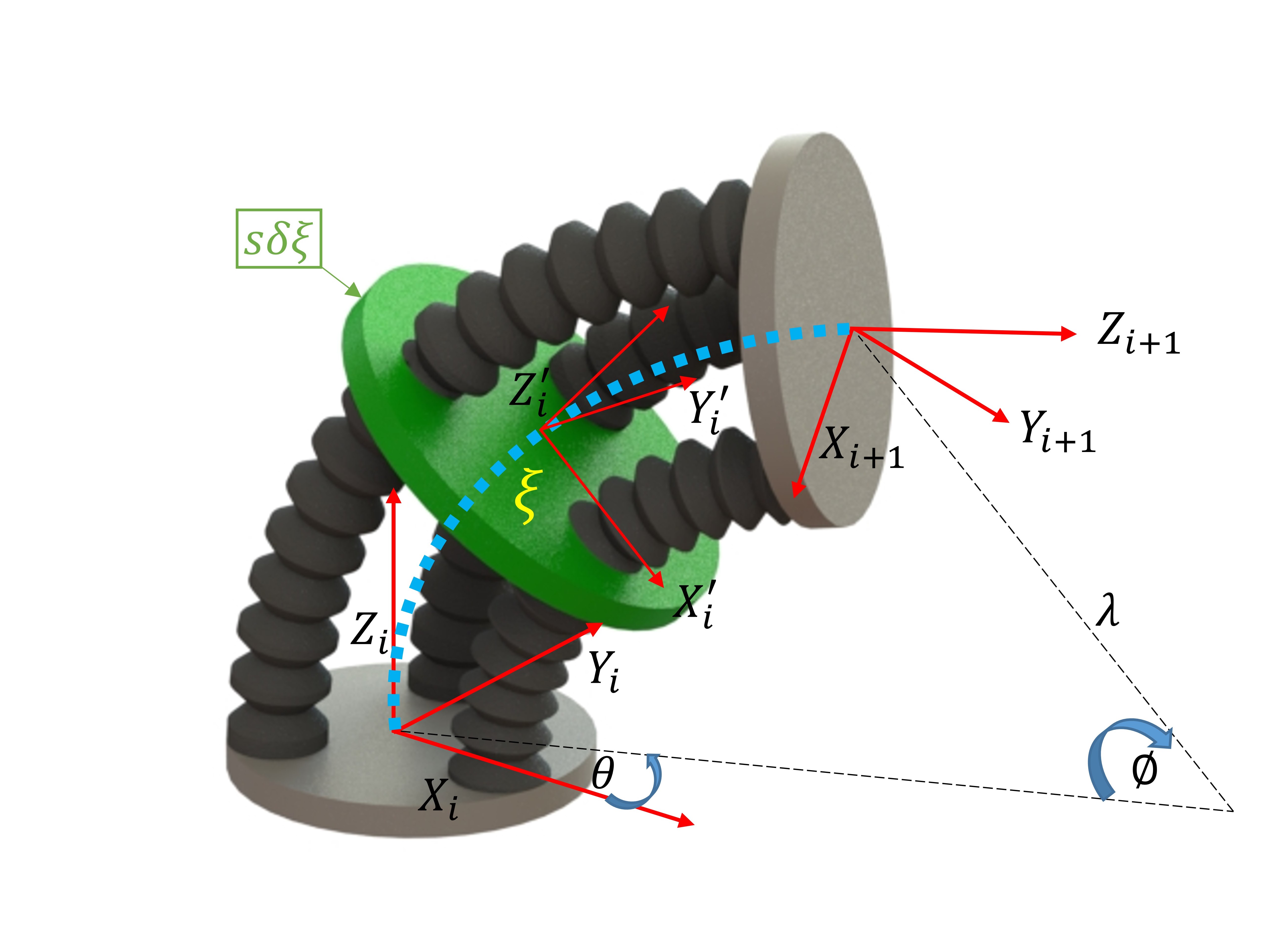}
    \caption{The one segment soft robotic arm consists of three extending actuators. The moving frame in the soft robot is shown with a green disk which is located at the $\xi$ position of the neutral axis. The thin slice of $s \delta \xi$ is used to derive the dynamic parameter of the system using integration along  the robot}
    \label{fig:robot}
\end{figure}

In this paper, we only focus on the single segment soft robot arm with three identical extending actuators separated by 120$^{\circ}$ (see Figure \ref{fig:robot} for the robot schematic diagram). The extending actuator has the nominal length of $L_{0}$, and the length of each actuator under actuator space input could be described as follows:
\begin{equation}
    L_{i}(t) = L_{0} + l_{i}(t)
\end{equation}
where $i \in [1,2,3]$ represent each actuator and $l_{i}(t)$ is the variable length of each actuator caused by actuator space input.

To derive the forward kinematics, the mapping between actuator space to configuration space, as well as the mapping between configuration space to Cartesian space should be derived.
The actuator space is defined as the variable length of each actuator $q = [l_{1}, l_{2},l_{3}]^{T} \in \mathbb{R}^{3}$. 
The configuration of the system is defined via following three parameters:
\begin{itemize}
    \item Radius of curvature, $\lambda \in (0, \infty)$
    \item Bending angle of the arc, $\phi \in [0, 2\pi)$
    \item Angle of the bending plane with respect to X-axis, $\theta \in [-\pi, \pi]$ (refer to Figure \ref{fig:robot} for the coordinate frame definition)
\end{itemize}

The mapping between actuator space and configuration space, which could be derived using the geometrical shape of the robot,  is described as follows:
\begin{equation}
\begin{aligned}
        \lambda(q) &= \frac{3L_{0}+l_{1}+l_{2}+l_{3}}
    {2\sqrt{{l_{1}}^2+{l_{2}}^2 + {l_{3}}^2 -l_{1}l_{2} - l_{2}l_{3} - l_{1}l_{3}}}\\
    \phi &= \frac{2\sqrt{{l_{1}}^2+{l_{2}}^2 + {l_{3}}^2 -l_{1}l_{2} - l_{2}l_{3} - l_{1}l_{3}}}{3r}\\
    \theta &= \arctan\left({\frac{\sqrt{3}(l_{3}-l_{2})}{l_{2} + l_{3} - 2l_{1}}}\right)
\end{aligned}
\label{eq:eq_1}
\end{equation}
\\
With the mapping between actuator space to configuration space, the robot position and orientation in Cartesian space should be calculated with an additional mapping. 
A scalar parameter $\xi \in [0, 1]$ is introduced to describe the virtual moving disk location along the robot from its base $\xi = 0$, to its tip $\xi = 1$. The moving disk is also shown in Figure \ref{fig:robot}. The position and orientation of the robot are described via a homogeneous transformation $T(\xi, q)$:
\begin{equation}
    T(\xi, q) =
    \begin{bmatrix}
    R(\xi, q) & P(\xi, q)\\
    0 & 1
    \end{bmatrix}
    \label{eq:eq_2}
\end{equation}
where $R(\xi, q) \in \mathbb{R}^3$ defines the orientation, and $P(\xi, q) \in \mathbb{R}$ defines the position of the robot. The element of the transformation matrix can be derived as follows:
\begin{equation}
    T(\xi,q) = \text{Rot}_{z}(\theta)\text{Trans}_{x}(\lambda)\text{Rot}_{y}(\xi \phi) \text{Trans}_{x}(-\lambda)\text{Rot}_{z}(-\theta)
\end{equation}

\subsection{Dynamic Modeling}
The dynamic modeling of the soft robot is derived via the Lagrangian Formulation. The complete dynamic equation of the robot which also includes the non-conservative terms is described as follows:
\begin{equation}
    M\Ddot{q}+C\Dot{q}+
    D\Dot{q} + Kq + G = \tau
    \label{eq: dynamic-compact}
\end{equation}
where $M \in \mathbb{R}^{3 \times 3}$ denotes the inertial matrix of the system, $C \in \mathbb{R}^{3 \times 3}$ defines the centrifugal and Coriolis force matrix, $D \in \mathbb{R}^{3 \times 3}$ represent the damping coefficient matrix, $K \in \mathbb{R}^{3 \times 3}$ is the stiffness coefficient matrix of the system, and $G \in \mathbb{R}^3$ is the gravitational force vector. The parameters are derived based on our prior work \cite{godage2016dynamics}. In what follows, we will derive the formulation of each dynamic parameter.
\subsubsection{Inertial Matrix, M}
The Inertial Matrix is defined as follows:
\begin{equation}
    M = M^{w} + M^{v}
    \label{eq: M-matrix}
\end{equation}
where $M^{w}$ and $M^{v}$ are associated with angular and linear dynamics of the robot. It has been assumed that the robot is consists of infinitesimally thin slices with a constant mass and uniform linear density, which has been shown in Figure \ref{fig:robot}.
The $M^{w}$, and $M^{v}$ is defined as follows:
\begin{equation}
\begin{aligned}
            M^{w}[j,k] &= I_{xx} \int_{\xi} \mathbb{T}_{2} (\frac{\partial R}{\partial q(j)})(\frac{\partial R}{\partial q(k)})\\
            M^{v}[j,k] &= m \int_{\xi} (\frac{\partial P}{\partial q(j)})(\frac{\partial P}{\partial q(k)})
\end{aligned}
\end{equation}
where $\mathbb{T}_{2}$ is defined as the operator which calculates the sum of the first two elements of the principal diagonal of $3 \times 3$ Matrix.  $I_{xx}$ is the moment of Inertia along the X-axis of the slice, $m$ is the mass of the system, and $j, k \in [1,2,3]$.
\subsubsection{Centrifugal and Coriolis force Matrix, C}
\begin{equation}
    \begin{aligned}
    C[k,j] &= \sum_{i=1}^{3} \Gamma_{ijk}(M) \Dot{q}(i)\\
    \Gamma_{ijk}(M) &= \frac{1}{2} \left( \frac{\partial M[k,j]}{\partial q(i)} + \frac{\partial M[k,i]}{\partial q(j)}  -\frac{\partial M[i,j]}{\partial q(k)}\right)
    \end{aligned}
    \label{eq: C-Matrix}
\end{equation}
\subsubsection{Gravitational Vector, G}
To derive the gravitational vector, the Jacobian of linear velocity in moving frame $J^{v} \in \mathbb{R}^{3 \times 3}$ for i-th column is derived as follows:
\begin{equation}
    J^{v}(i) = R^{T} \frac{\partial P}{\partial q(i)}
\end{equation}
where $i \in [1,2,3]$. Now that the Jacobian of linear velocities is defined, the i-th element of gravitational vector could be derived as follows:
\begin{equation}
    G(i) = m \int_{\xi} {J^{v}}^{T}(i)R^{T}G_{v}
\end{equation}
where $i$ and $G_{v}$ is defined as $i \in [1, 2, 3]$ and $G_{v} = [0,0,g]^{T}$ respectively.
\subsubsection{Stiffness Matrix, K}
At this stage, the stiffness of each actuator has been considered linear and independent of each other. Additionally, rotational stiffness for this system was not considered. Therefore, the stiffness matrix can be written in a diagonal matrix as  $K = \text{diag}([K_{1},K_{2},K_{3}])$, where each element can be experimentally calibrated prior to the controller implementation.
\subsubsection{Damping Coefficient Matrix, D}
Similar to the stiffness matrix, damping coefficient matrix can also be defined as a diagonal matrix, $D = \text{diag}([D_{1},D_{2},D_{3}])$, with each element pre-calibrated prior to the experiment.
\subsubsection{Hysteresis effect}
Pneumatically actuated actuators demonstrate an inherent hysteresis nature in their dynamics \cite{godage2018dynamic}. This indicates that the system presents a different profile in loading and unloading phases. The hysteresis phenomenon can be modeled via Preisach, Prandtl-Ishlinskii, or Max-well-slip model \cite{thai2021design}. In this paper, we consider the Bouc-Wen model for its unique differential modeling that could decrease the complexity in both implementation and computation\cite{thai2021design}. The Bouc-Wen model can be written as
\begin{equation}
    \dot{h}(i) = q(i)\left[\alpha_{h} - \left\{\beta_{h} \text{sgn}(\dot{q}(i)h(i)) + \gamma_{h} \right\}|h(i)|  \right]
\end{equation}

With the hysteresis effect in the practical scenarios, the system dynamic model \eqref{eq: dynamic-compact}  can be updated  as follows:
\begin{equation}
    M\Ddot{q}+C\Dot{q}+
    D\Dot{q} + Kq + G + H = \tau
\end{equation}
where $H = [h(1),h(2),h(3)]^{{T}}$ is the hysteresis effect of each actuator. 
Now that the dynamic model of the system has been briefly reviewed. In the next part, we focus on the control of the soft robotic arm  under various operation scenarios. 
\section{Soft Robot Control}
\label{sec:controlcomparison}
Soft robots are considered as complex dynamical systems for their continuum and nonlinear nature. In these systems, deriving the exact dynamic model is a time-consuming and challenging procedure, and at best could result in a sufficiently accurate model but cannot handle the practical applications caused by parametric uncertainties or any potential unmodelled terms. To overcome the aforementioned challenges, a proper controller should be implemented to compensate model inaccuracy and uncertainties while maintaining the closed-loop stability of the system. In this section, we investigate several conventional robot controllers to achieve the desired task space control. The performance of each controller is discussed, and their shortcomings are presented via simulation studies.  

\subsection{Kinematic Control}
Kinematic controller is the simplest form of the control scheme. The kinematic controller is described as follows:
\begin{equation}
    \tau = K_{kin}e_{k}, \quad e_{k} = q_{d}-q
\end{equation}
where $K_{kin} \in \mathbb{R}^{3 \times 3}$ is a proportional diagonal matrix, and $e_{k}$ is defined as a difference between the desired actuator space vector $q_{d}$  and the current actuator space vector $q$ of the robot.
However, kinematic control is unable to capture the dynamic behavior of the robot, especially during the transient state. Moreover, the kinematic model control completely relies on the perfect identification of $K_{kin}$, which might not be accurate in real-world applications. 
\subsection{PD + Feedback Linearization}
To overcome the shortcoming of the kinematic controller, especially the transient state performance,  a dynamic proportional and derivative (PD)  Feedback linearization (FL) controller \cite{spong2008robot} can be used for soft robot control. This controller is designed based on the exact cancellation of the dynamic part to transform the system into a linear system. Then, by using proportional and derivative gains, the controller is designed in a way to achieve the desired  transient response of the robot. The  PD  Feedback linearization controller is defined as follows:
\begin{equation}
    \begin{aligned}
    \tau &= \alpha \tau^{\prime} + \beta,\quad \tau^{\prime} = \Ddot{q}_{d} - K_{d}\Dot{\tilde{q}} - K_{p}{\tilde{q}}\\
    \alpha &= M, \quad \beta = C\Dot{q} + D\Dot{q} + Kq + G
    \end{aligned}
    \label{eq:PDC}
\end{equation}
where $\beta$ is the feedback linearization term that cancels out the nonlinear dynamics of the system, and $\tau^{\prime}$ would be designed in a way to make the system follow the desired trajectory with prescribed performance defined by the PD controller. $\tilde{q}$ is defined as the difference between the current state and the desired state of the robot $\tilde{q} = q - q_{d}$. Now, substituting \eqref{eq:PDC} into the dynamic equation of the robot \eqref{eq: dynamic-compact}, the closed-loop system would become as follows:
\begin{equation}
    \ddot{\tilde{q}} + K_{d}\dot{\tilde{q}} + K_{p}\tilde{q} = 0
\label{eq:pdeq}
\end{equation}

By choosing $K_{p}$ and $K_{d}$ as diagonal matrices as $K_{p}(i) = (\omega_{i})^{2}$ and $K_{d}(i)=2\omega_{i}$, the system transforms to a decoupled closed-loop system with a critically damped linear response in each actuator. The natural frequency of each actuator is defined as $\omega_{i}$, which determines the decay rate of the error, as well as the response time.
Based on \cite{spong2008robot} and Lassalle theorem we can conclude that the system is asymptotically stable.

The PD+FL approach has its own shortcomings. For instance, if the system dynamics are not derived carefully, or the identification parameters are not accurate, the model would have uncertainty in its parameters. Thus, the asymptotic tracking might not be possible and the system could have steady-state error.  To compensate these uncertainties, robust or adaptive terms should be included in the control scheme design.

\subsection{Passivity Based Control}
Now in this section, we are considering modeling uncertainties, which mainly come from an inaccurate experimental calibration, or the wear of materials due to usage of the system.
The feedback linearization methods rely on the exact cancellation of all the system's nonlinearities. The passivity control scheme relies on the passivity or the skew-symmetry property of Euler-Lagrange equations. The great advantage of passivity control comes in a situation where the robot has uncertainties in the model\cite{spong2008robot}.  The passivity controller itself has an additional term that works as a gain for a filtered error, which helps the controller to compensate unmodeled dynamics. 
In the passivity based control, the controller is defined as   
\begin{equation}
\begin{aligned}
&v =\dot{q}^{d}-\Lambda \tilde{q}, 
a =\dot{v}=\ddot{q}^{d}-\Lambda \dot{\tilde{q}},
r =\dot{q}-v=\dot{\tilde{q}}+\Lambda \tilde{q}\\
&\tau = M(q)a + C(q,\dot{q})v + G(q) + Kq + Dv - K_{G}r
\end{aligned}
\end{equation}
where $\Lambda \in \mathbb{R}^{3 \times 3}$, $K_{G} \in \mathbb{R}^{3 \times 3}$ are diagonal positive constant matrices, which are defined in Table. \ref{tab:tab1} during the simulation.

\subsection{Passivity Based Adaptive Control}
As discussed before, the system model parameters could be uncertain or even cannot be accurately identified. For example, the stiffness and damping coefficient matrices are typically experimentally identified to minimize the least square error. However, these parameters could potentially change with respect to time and temperature. In this section, we aim to overcome this uncertainty by using the passivity based adaptive control, which can be defined as follows: 
\begin{equation}
    \tau=\hat{M}(q) a+\hat{C}(q, \dot{q}) v+\hat{G}(q)-K_{G} r + \hat{K} q + \hat{D} v
    \label{eq:passiveAdapt}
\end{equation}
where $\hat{[\quad]}$ indicates the estimated parameters. The general control signal could also be described in the regressor form as follows:
\begin{equation}
    \tau=Y(q, \dot{q}, a, v) \hat{\theta}_{p}-K_{G} r
    \label{eq:AdaptPass}
\end{equation}
where $Y$ is the regressor matrix and $\theta_{p}$ is the estimated.
Based on \cite{spong2008robot}, the adaptive rule could be designed as follows:
\begin{equation}
    \dot{\hat{\theta}}_{p}=-\Gamma^{-1} Y^{T}(q, \dot{q}, a, v) r
\end{equation}

It should be noted that in our experiment, only the stiffness and damping were assumed to have uncertainty, and other terms implemented in the control signal using Eq. \eqref{eq:passiveAdapt}. In the following equation, the regressor and the estimated parameter are defined.
\begin{equation}
\begin{aligned}
    Y &= [\text{diag}([q(1),q(2),q(3)]),\text{diag}([v(1),v(2),v(3)])]\\
    \hat{\theta}_{p} &= [\hat{K}_{1},\hat{K}_{2},\hat{K}_{3},\hat{D}_{1},\hat{D}_{2},\hat{D}_{3}]^{T},
\end{aligned}
\end{equation}
\subsection{Hysteresis Compensation }
To compensate the hysteresis effect, a feasible solution is to add the identified hysteresis dynamics to the feedback linearization loop. For instance, if the hysteresis of the system could be identified with good accuracy, the related terms would be able to compensate those effects, which is true in our prior studies \cite{godage2018dynamic}. Thus, the feedback linearization term can be re-written as

\begin{equation}
    \beta_m = C\dot{q}+D\dot{q}+Kq+G+H
    \label{eq:hystCompensate}
\end{equation}
\section{Controller Implementation and Comparison}
\label{sec:controlimplementation}

In this section, the control approaches in the prior section will be implemented to track the desired trajectory and their performance will be compared with each other. First, the dynamic controllers are compared with the kinematic control approach. In the next part, the effect of parametric uncertainty will be investigated. Then, hysteresis is added to the model, and the performance of controllers for compensating the effect of hysteresis is studied in detail. The system parameters  are shown in Table. \ref{tab:tab1}, which are chosen based on our previous works \cite{godage2016dynamics,godage2018dynamic}. The desired path is described as follows:
\begin{equation}
    X_{d} = 0.1 sin(3t), \quad
    Y_{d} = 0.1 cos(3t), \quad
    Z_{d} = 0.147
    \label{eq:trajectoryTrack}
\end{equation}

\begin{table}[htb!]
    \centering
        \begin{tabular}{|c|c|c|c|c|}
        \hline
             Parameter [Unit] & $K(i)$ $\left[ \frac{N}{m} \right]$ & $D(i)$ $\left[ \frac{N.s}{m}\right]$ & $m$ [$kg$] & $L_0$ [$m$]\\
        \hline
             Value & 1700 & 110 & 0.13 & 0.15\\ 
        \hline
            $g$ $\left[ \frac{m}{s^{2}}\right]$& $r$ [$m$] & $\alpha_{h}(i)$ & $\beta_{h}(i)$ & $\gamma_{h}(i)$\\
        \hline
            -9.81 & 0.0125 & 23.705 & 1.7267 & -42.593\\
        \hline
            $K_{G}(i)$ & $K_{p}(i)$ & $K_{d}(i)$ &$\Lambda(i)$ & $\Gamma^{-1}(i)$\\
        \hline
            10 & $10^{4}$ & 200 & $10^{2}$ & $10^{5}$\\
        \hline
        \end{tabular}
    \caption{The Parameters for Simulation}
    \label{tab:tab1}
\end{table}
\subsection{Kinematic vs Dynamic Controllers Without  Uncertainties}

\begin{figure}[bth]
     \centering
    \includegraphics[width=0.38\textwidth]{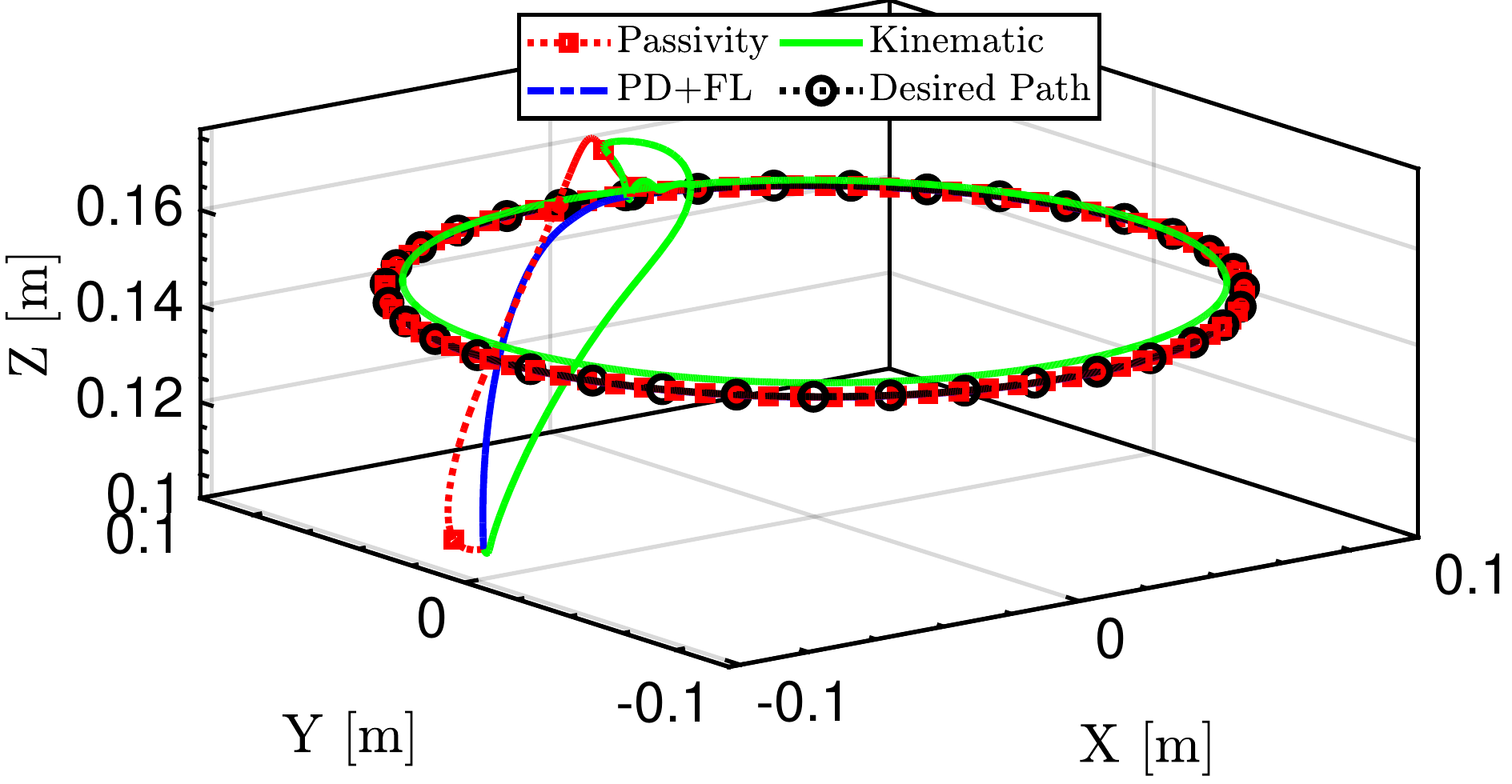}
    \label{fig:XYZ kin}
     \caption{Path tracking performance comparison of kinematic vs dynamic controllers in task space. The initial position of the robot end-effector in Cartesian space is $[-0.0990,-0.0017,0.1067]^{T}$.}
     \label{fig:PathTrackKD}
\end{figure}

The tracking results of both kinematic and dynamic controllers are presented in Figure \ref{fig:PathTrackKD}.
It should be noted that in this section we assume that the model is accurate and there are no uncertainties.  The kinematic controller has shown an obvious overshoot and increase in the error magnitude at the beginning of the tracking task, while PD+FL controller reaches the steady-state error with a smooth damped response. The passivity-based controller demonstrated comparable performance with respect to PD+FL controller.  It should be noted that PD+FL controller was designed to achieve a critically-damped response in actuator space. However, the actuator space critical damping behavior can not guarantee the critical damping in the task space.
To achieve the critical damping in the task space control, the robot dynamics parameter should be described in Cartesian space, and the controller parameters should be redesigned accordingly. 

\begin{figure}[tbh]
    \centering
    \includegraphics[width=0.38\textwidth]{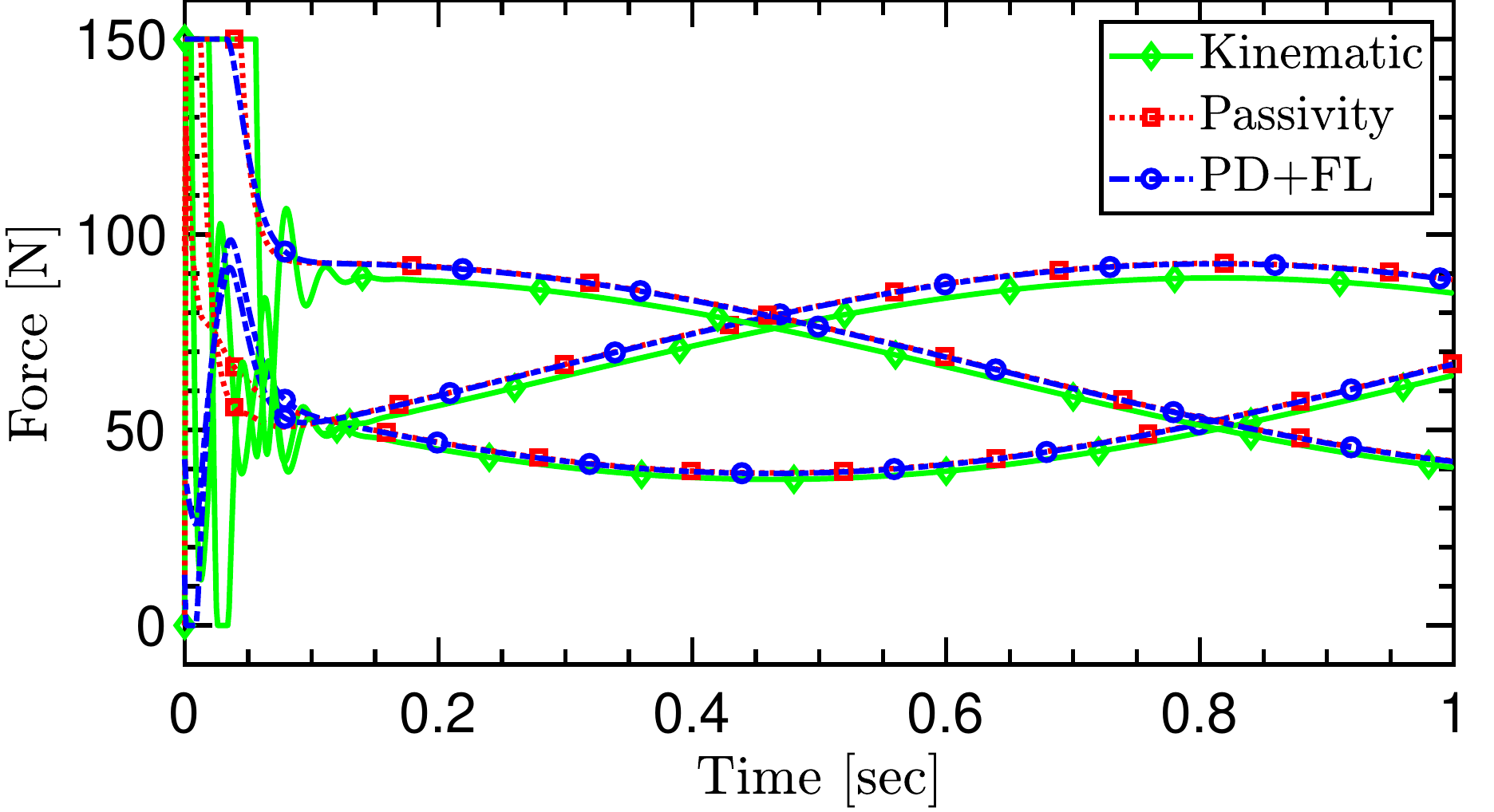}
    \caption{Control effort comparison between kinematic and dynamic controllers. 
    }
    \label{fig:cont comp kin}
\end{figure}

The magnitude of the calculated control signal,  which is depicted in Figure \ref{fig:cont comp kin}, in the kinematic controller at the beginning of the simulation is significantly higher than the dynamic approaches, despite it has been saturated during simulation. Since the robot could not withstand significantly high pressures in the practical scenarios, the saturation block should be implemented on control output. As can be seen in Figure \ref{fig:cont comp kin}, almost every actuator has been saturated in kinematic control.  This result comes from a fact that the kinematic control scheme doesn't take into account the dynamic effects of the system. The dynamic controllers demonstrated better control command in the transient state. 

\begin{figure*}[bth]
     \centering
    \begin{subfigure}[b]{0.3\textwidth}
         \centering
         \includegraphics[width=\textwidth]{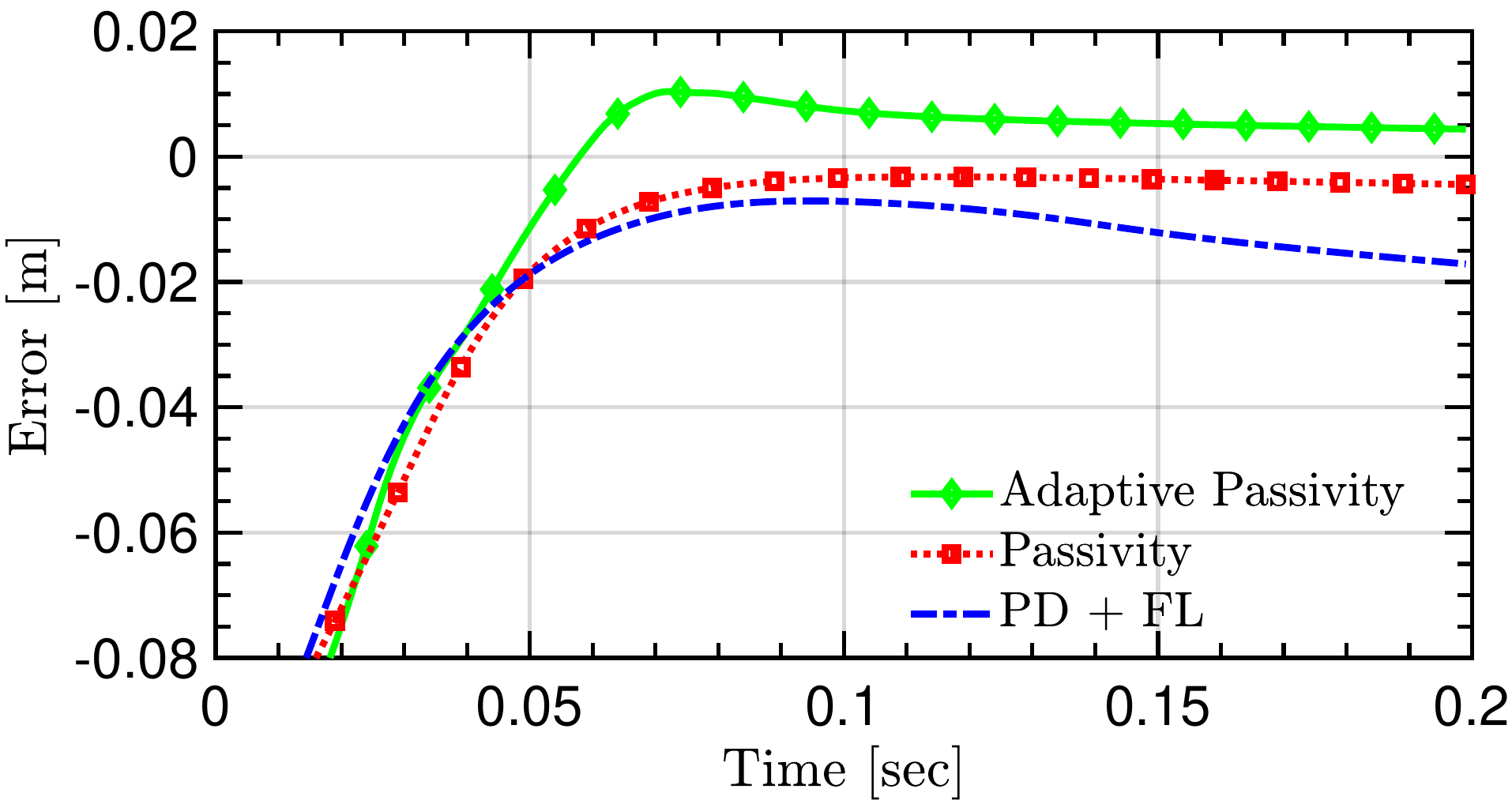}
         \caption{}
         \label{fig:Par d}
     \end{subfigure}
     \hfill
     \begin{subfigure}[b]{0.3\textwidth}
         \centering
         \includegraphics[width=\textwidth]{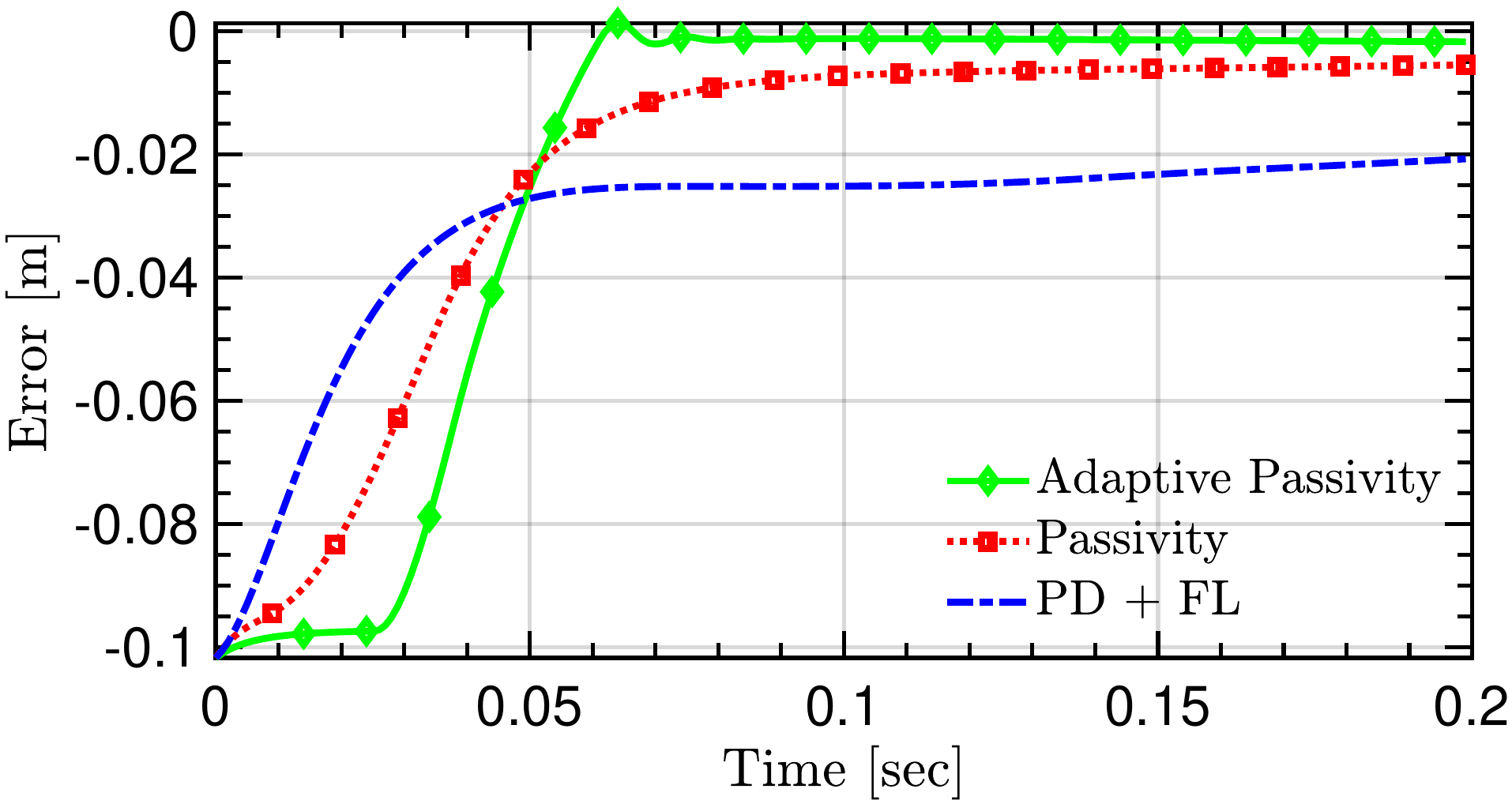}
         \caption{}
         \label{fig:Par e}
     \end{subfigure}
     \hfill
     \begin{subfigure}[b]{0.3\textwidth}
         \centering
         \includegraphics[width=\textwidth]{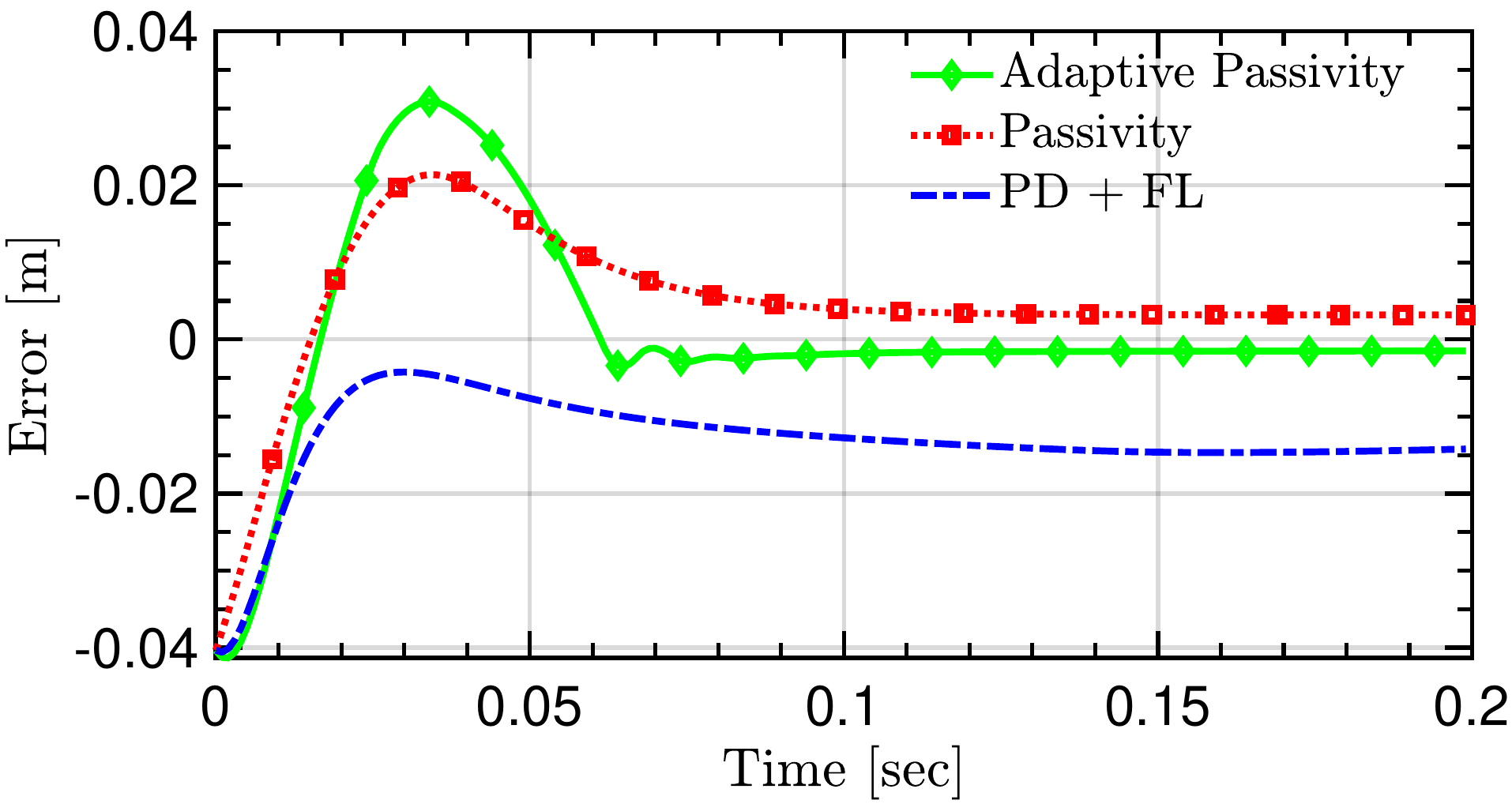}
         \caption{}
         \label{fig:Par f}
     \end{subfigure}
     \begin{subfigure}[b]{0.3\textwidth}
         \centering
         \includegraphics[width=\textwidth]{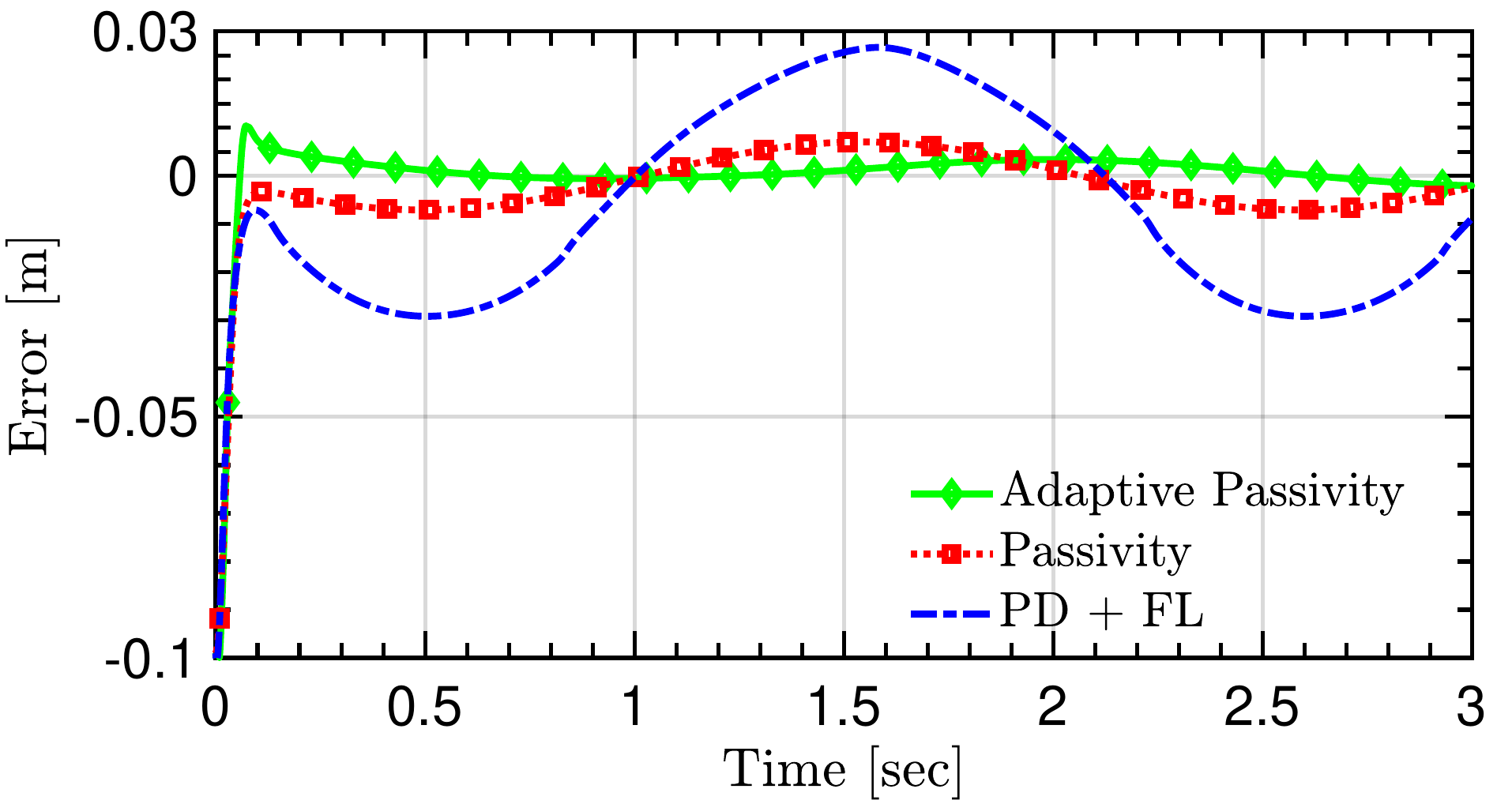}
         \caption{}
         \label{fig:Par a}
     \end{subfigure}
     \hfill
     \begin{subfigure}[b]{0.3\textwidth}
         \centering
         \includegraphics[width=\textwidth]{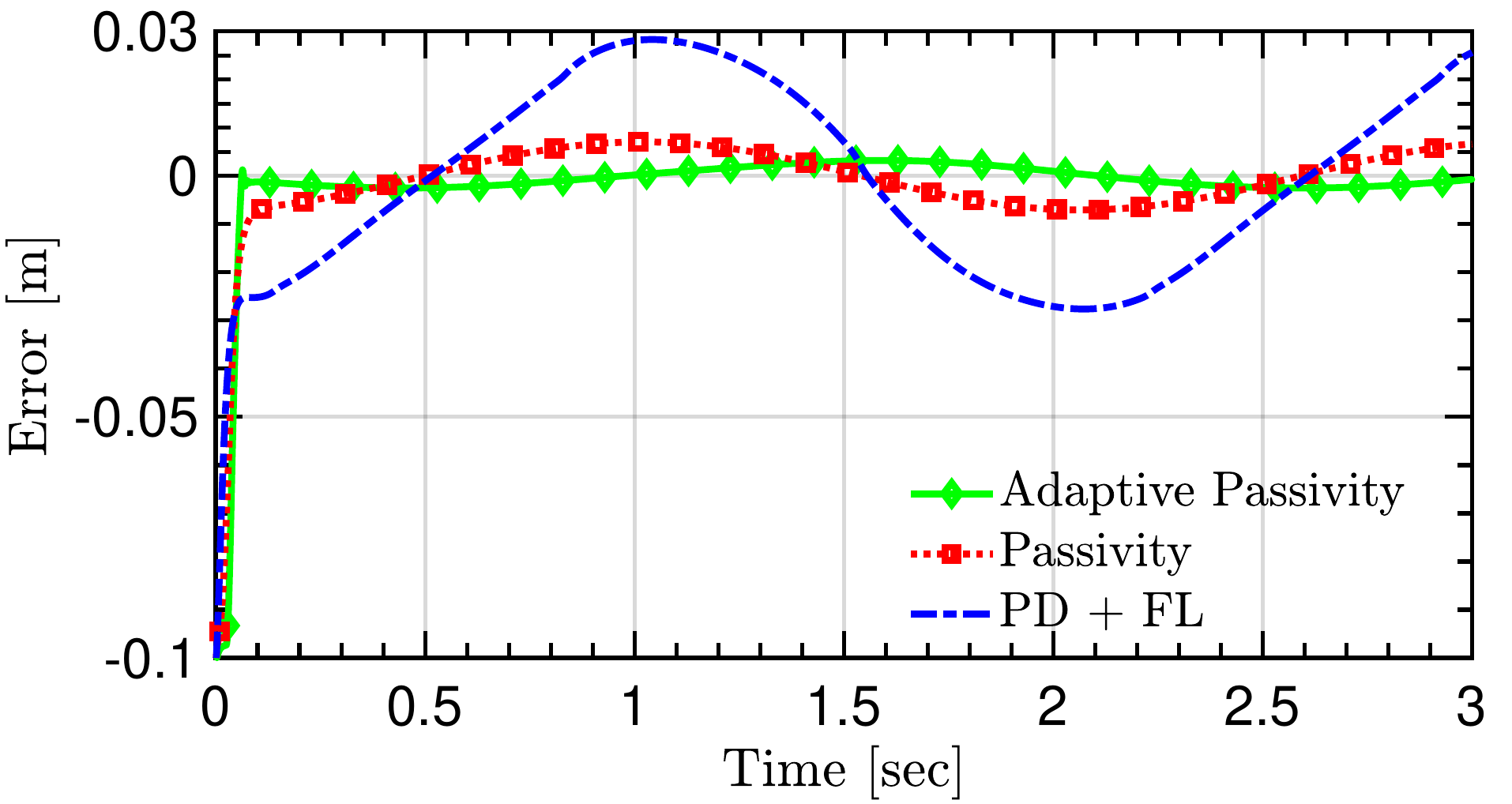}
         \caption{}
         \label{fig:Par b}
     \end{subfigure}
     \hfill
     \begin{subfigure}[b]{0.3\textwidth}
         \centering
         \includegraphics[width=\textwidth]{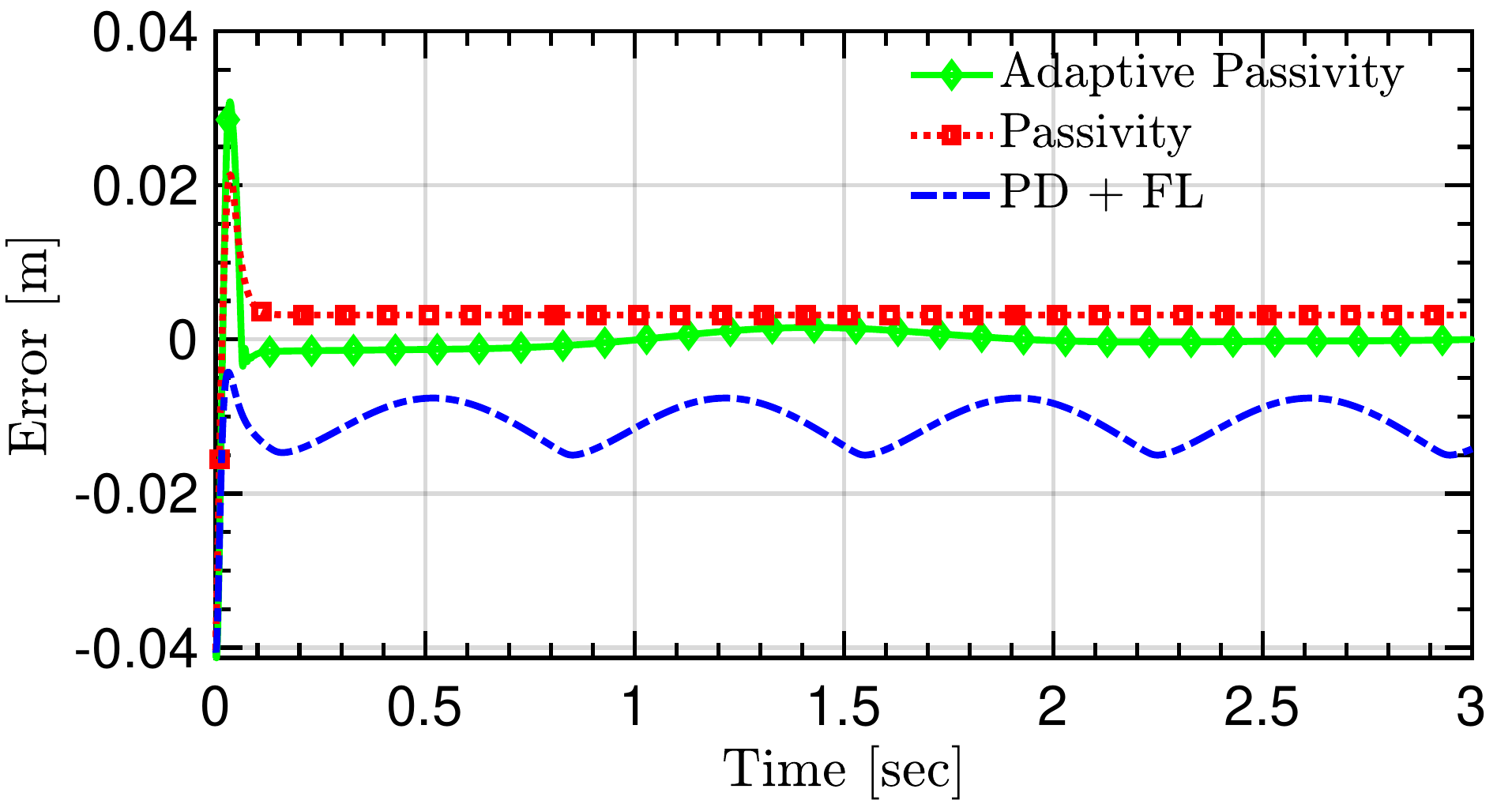}
         \caption{}
         \label{fig:Par c}
     \end{subfigure}
        \caption{Tracking performance comparison of PD+FL, passivity control, and adaptive passivity  control in the presence of parametric uncertainty. Steady-state response is depicted in (a) X-direction, (b) Y-direction, and (c) Z-direction. The transient response is also shown in sub-figure (d), (e), and (f) for X, Y, Z direction respectively.}
        \label{ErrorTrackParam}
\end{figure*}

\subsection{Robot Control with Model Uncertainties}
As we discussed earlier, the soft robot model parameters could be uncertain due to the inaccurate system identification or wear in the system. The control scheme should be designed in a way to compensate these uncertainties. In this section, the conventional dynamic controllers (PD+FL and Passivity) are compared with the adaptive passivity-based controller. We only consider the uncertainties in stiffness and damping coefficient matrices since both parameters require identification prior to the practical control \cite{godage2018dynamic}. These uncertainties are defined as $ K_{e} = \text{diag}([1020,1020,1020])$ and $ D_{e} = \text{diag}([77,77,77])$  in the simulations, respectively.

As can be seen in Figure \ref{fig:Par d}, \ref{fig:Par e}, and \ref{fig:Par f}, the adaptive passivity controller could track the desire path, while PD+FL didn't even get close to the desired trajectory because its performance highly depends on exact cancellation of the nonlinear dynamics. Since the parameters are uncertain, the controller may not be able to achieve the desired performance. In comparison, the passivity-based controller performed better compared to PD+FL controller since it does not completely rely on perfect cancellation of the nonlinear dynamics. In steady-state responses, which are shown in Figure \ref{fig:Par a}, \ref{fig:Par b}, and \ref{fig:Par c}, we can see that the adaptive passivity controller present the smallest error, and the passivity-based controller performed better than the PD+FL.

\begin{figure}[tbh!]
    \centering
    \includegraphics[width=0.38\textwidth]{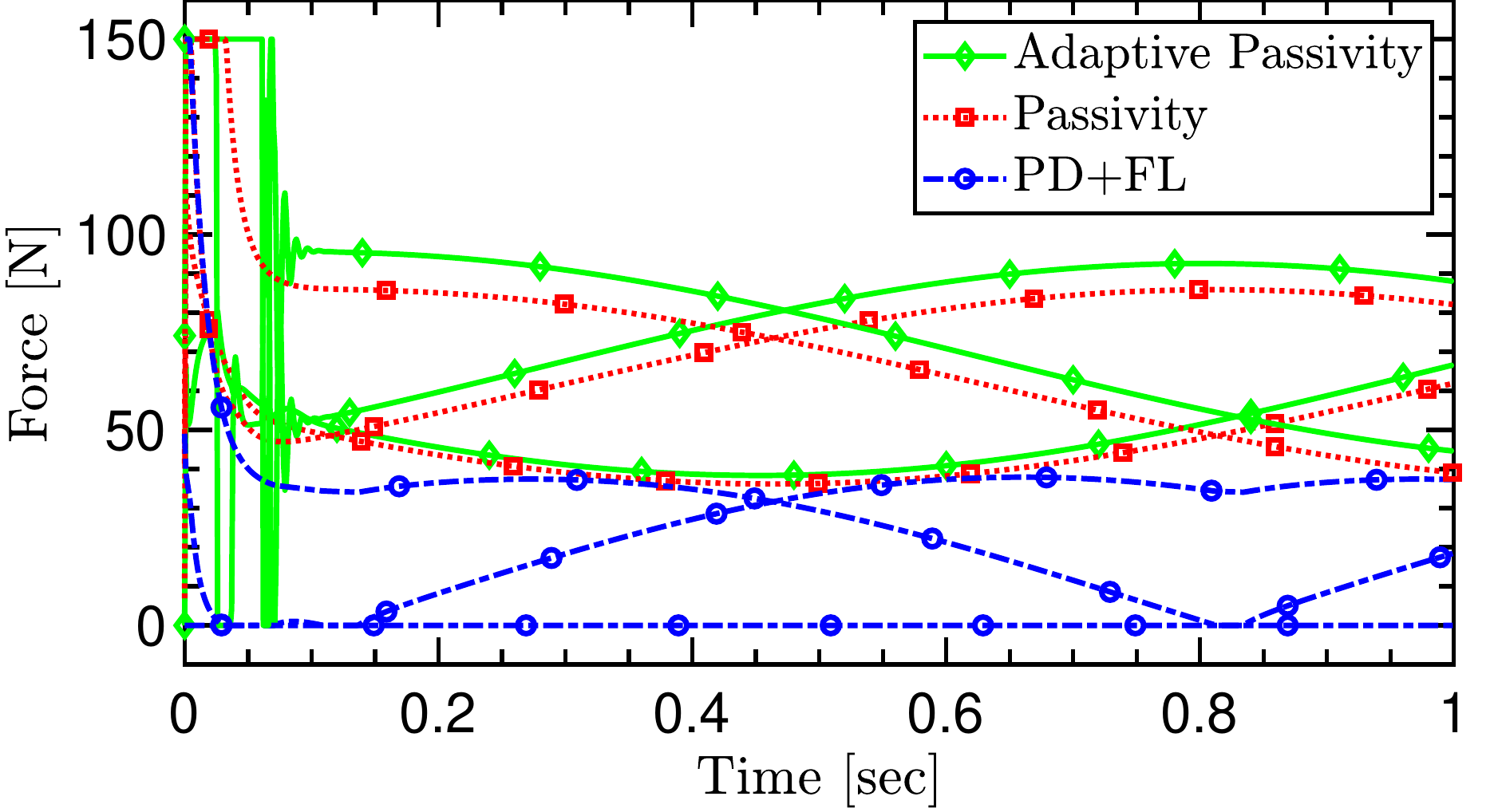}
    \caption{Control effort comparison between adaptive and conventional dynamic controllers (PD+FL and Passivity) in the presence of parametric uncertainty. 
    }
    \label{fig:compcss}
\end{figure}
Although the adaptive-passivity controller presents a higher magnitude force at the transient response of the simulation, which is depicted in transient  part of  Figure \ref{fig:compcss}, this should not be a problem because we can implement saturation at the input channel to the robot. Moreover, the duration of high magnitude input is only near 0.07$s$. In addition, the adaptive-passivity  controller could be designed with lower adaptation gain, which could result in more smooth control output with the limitation of slower reach time. In the steady-state response part, which is shown in Figure \ref{fig:compcss}, we can see that the adaptive-passivity controller generates a higher magnitude force in comparison with other methods. This result is associated with the adaptive terms that have tried to compensate the parametric uncertainties in the system. The PD+FL has the worst performance and the generated signal is not even valid in the non-positive section. This would result in even worse performance in tracking a trajectory when the pressure should decrease. 

\begin{figure}[tbh]
     \centering
         \centering
         \includegraphics[width=0.38\textwidth]{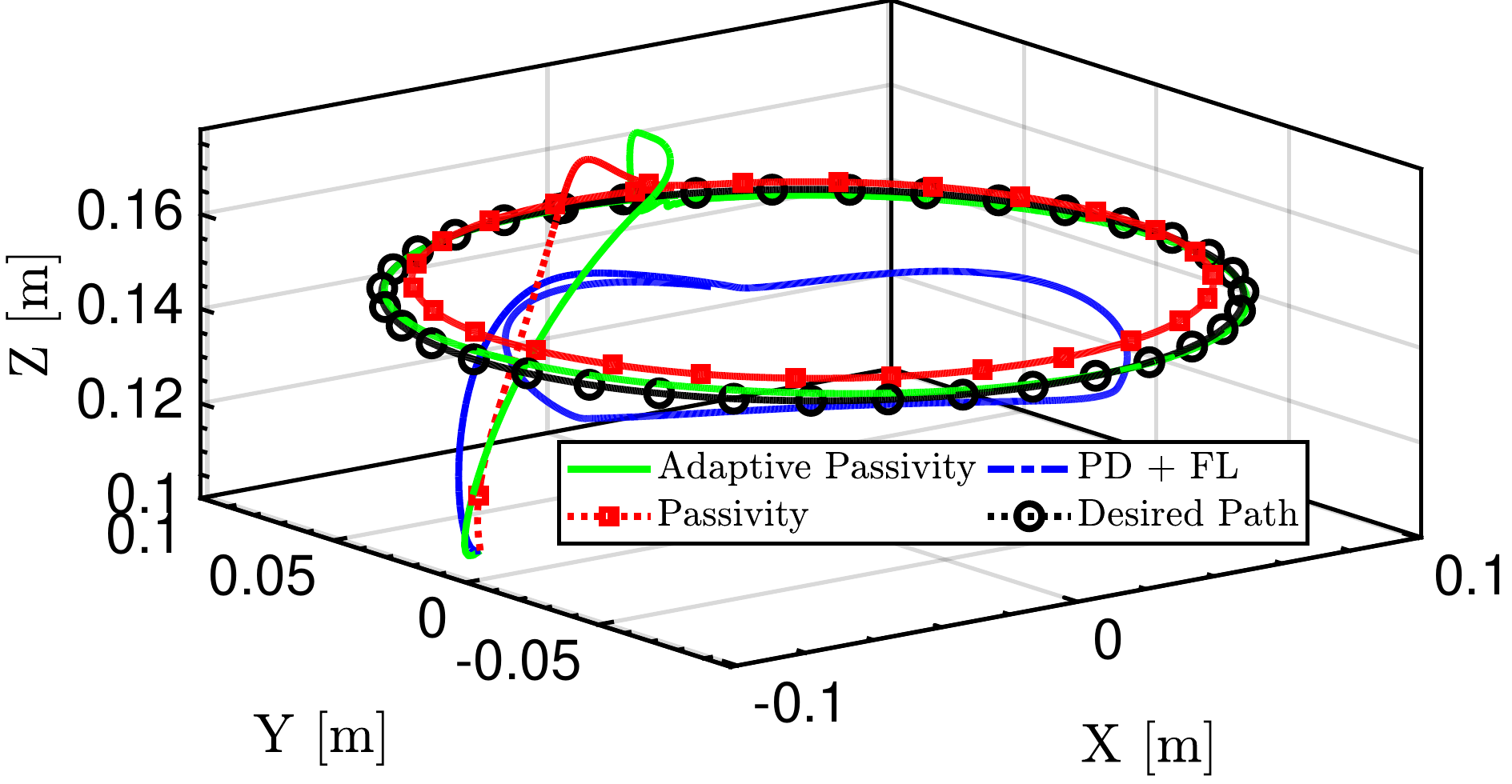}
     \caption{The tracking performance of different controllers in the presence of parametric uncertainty in Cartesian's space.  }
     \label{fig: PathTrackParam}
\end{figure}

With the above-mentioned arguments, the task space trajectory following of different controllers is shown in Figure \ref{fig: PathTrackParam}. As can be seen, the passivity controller showed a smooth movement, while the PD+FL have a higher error in tracking the desired path. The movement of the adaptive-passivity at the beginning of a simulation is not smooth, however, after some transient time, it could track the desired path with the best accuracy than other approaches.
\subsection{Hysteresis Compensation}
In this section, we want to validate the controller performance in soft robot hysteresis compensation for trajectory tracking based on Eq. \eqref{eq:trajectoryTrack}. We have assumed that the parameters have been modeled and identified accurately, and the only uncertainty/unmodelled dynamics is hysteresis effects. To address hysteresis effects we have compared the performance of mentioned approaches with the ideal controller that compensates the hysteresis effect with the term introduced in Eq. \eqref{eq:hystCompensate}. 

\begin{figure}[tbh]
    \centering
    \includegraphics[width = 0.38\textwidth]{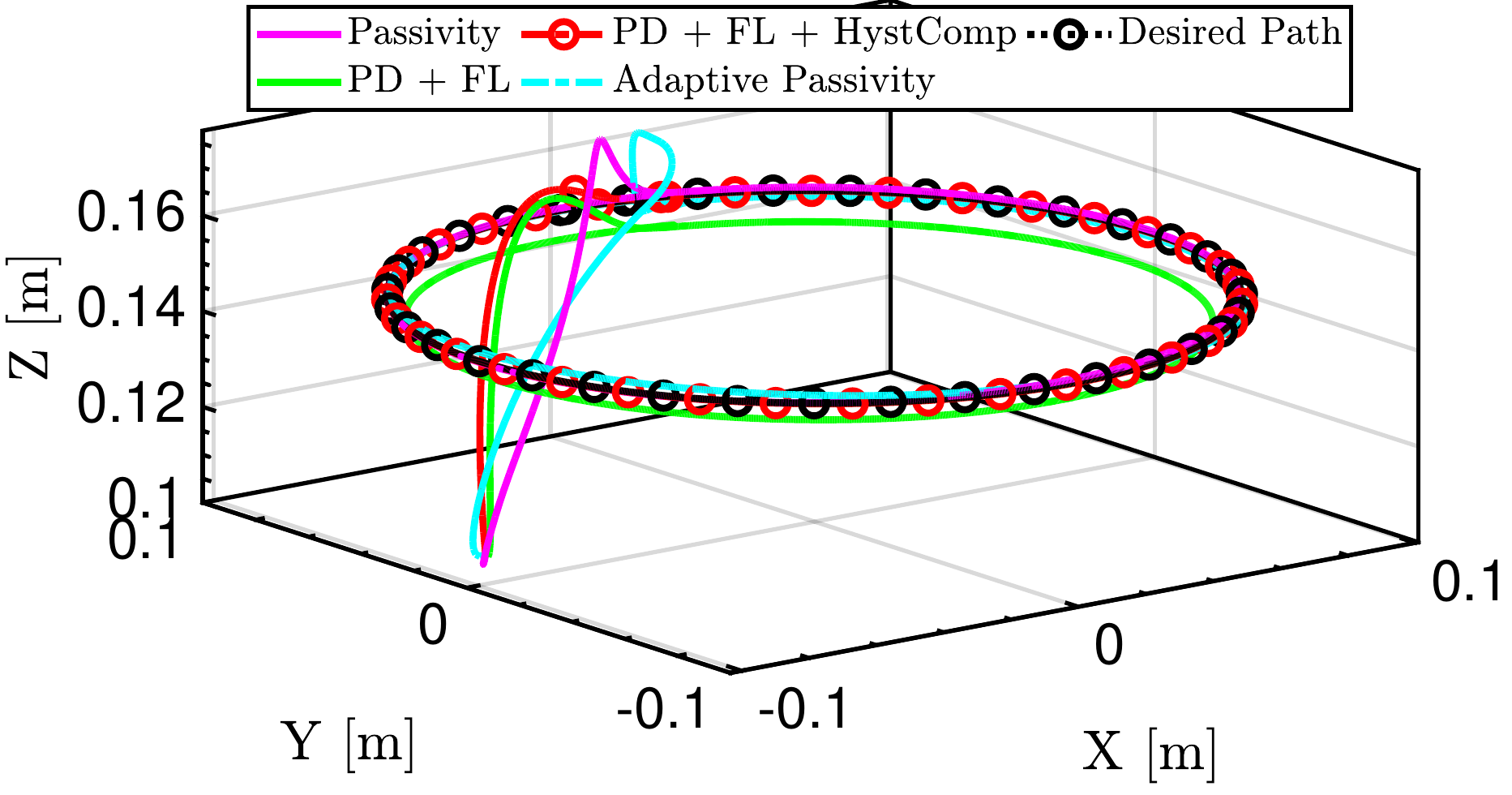}
    \caption{The tracking performance of different controllers in the presence of hysteresis in the system.}
    \label{fig:HysteresisCompTrack}
\end{figure}

As we can see in Figure \ref{fig:HysteresisCompTrack}, the PD+FL has the worst performance in tracking the desired path in comparison with other approaches. The passivity-based controller and adaptive passivity, have maintained their performance in the presence of hysteresis; however, it does not compensate the hysteresis completely. As can be inferred from Figure \ref{fig:HysteresisCompTrack}, if we can estimate the model accurately, the simple controller (PD+FL+ hysteresis compensation) can achieve the best tracking response. However, if this is not the case,  the adaptive-passivity controller demonstrated good performance despite of the parameter uncertainty and the presence of hysteresis effects. 

\section{Considerations for Practical Implementation}
\label{sec:practicalscenario}
In the previous section, the advantage and disadvantages of various control approaches have been discussed, and it has been shown that the adaptive-passivity based control outperforms the other approaches when uncertainties exist in the system. In this section, we consider additional challenges that will occur in the practical scenario, such as sensor noise, and the absence of velocity measurement of linear actuator. Moreover,  the prior modeling uncertainties and hysteresis  are considered in the simulation as well. 

\subsection{High Gain Observer}
To address the lack of velocity measurement, we propose to develop a high gain observer here\cite{Khalil:1173048}. By defining $x_{1} = q$, $x_{2} = \dot{q}$, and $x = \left[ {x_{1}}^{T},{x_{2}}^{T} \right]^{T}$ the dynamic equation of the robot can be described as follows: 
\begin{equation}
\begin{aligned}
    &\dot{x}_{1} = x_{2}, \quad \dot{x}_{2}= \Theta(x,\tau)\\
    &\Theta(x, \tau) = \text{inv}(M)[\tau - C\dot{x}_{2} - D\dot{x}_{2} - K{x}_{1} - G(x_{1})]
\end{aligned}
\end{equation}

In the situations where only the measurement of $x_{1}$ is available, the high gain observer can be used to estimate $x_{2}$ as follows:
\begin{equation}
    \dot{\hat{x}}_{1} = \hat{x}_{2} + g_{h1}(y-\hat{x}_{1}), \quad \dot{\hat{x}}_{2} = \Theta_{0}(\hat{x},\tau)+g_{h2}(y-\hat{x}_{1})
\end{equation}
where $y = x_{1} = q$, and $\Theta_{0}$ is the nominal plant. With that in mind, the estimation error is described as $\tilde{x} = x - \hat{x}$.
The estimation error derivative can be calculated as follows:
\begin{equation}
\dot{\tilde{x}}_{1}=-g_{h1} \tilde{x}_{1}+\tilde{x}_{2}, \quad \dot{\tilde{x}}_{2}=-g_{h2} \tilde{x}_{1}+\delta(x, \tau)
\end{equation}
where $\delta(x,\tau) = \Theta(x,\tau) - \Theta_{0}(\hat{x},\tau)$ is the disturbance term. To reject the disturbance term, the observer gains should be designed in way that $\text{lim}_{t \rightarrow \infty}\tilde{x}(t)=0$. This could not be achieved when the disturbance is present; However, by defining the observer gains as $g_{h2}(i) \gg g_{h1}(i) \gg 1 $, and the effects of the disturbance would be minimized\cite{Khalil:1173048}. 
\subsection{Robust Adaptive Term - Sigma Modification}
When the bounded disturbance is present in the system, the adaptive law should be modified to compensate those effects. Implementing robust adaptive terms could decrease the effects of noise and disturbances in the system. By using sigma-modification, the adaptation would be performed robustly and the parameters would not drift\cite{Lavretsky2013}. In addition, the higher adaptation gain could be used, while minimizing the effect of sensor noise in the control output. The robust adaptation rule with sigma modification terms\cite{Lavretsky2013} is described as follows:
\begin{equation}
    \dot{\hat{\theta}}_{p}=-\Gamma^{-1} Y^{T}(q, \dot{q}, a, v) r - \text{diag}([\sigma_{1},\sigma_{2},\sigma_{3}])\hat{\theta}_{p}
\end{equation}
Where $\sigma_{i}$ represent the sigma-modification constants.
\subsection{Simulation Results Considering Practical Scenarios}
\begin{figure}[htb]
    \centering
    \includegraphics[width=0.38\textwidth]{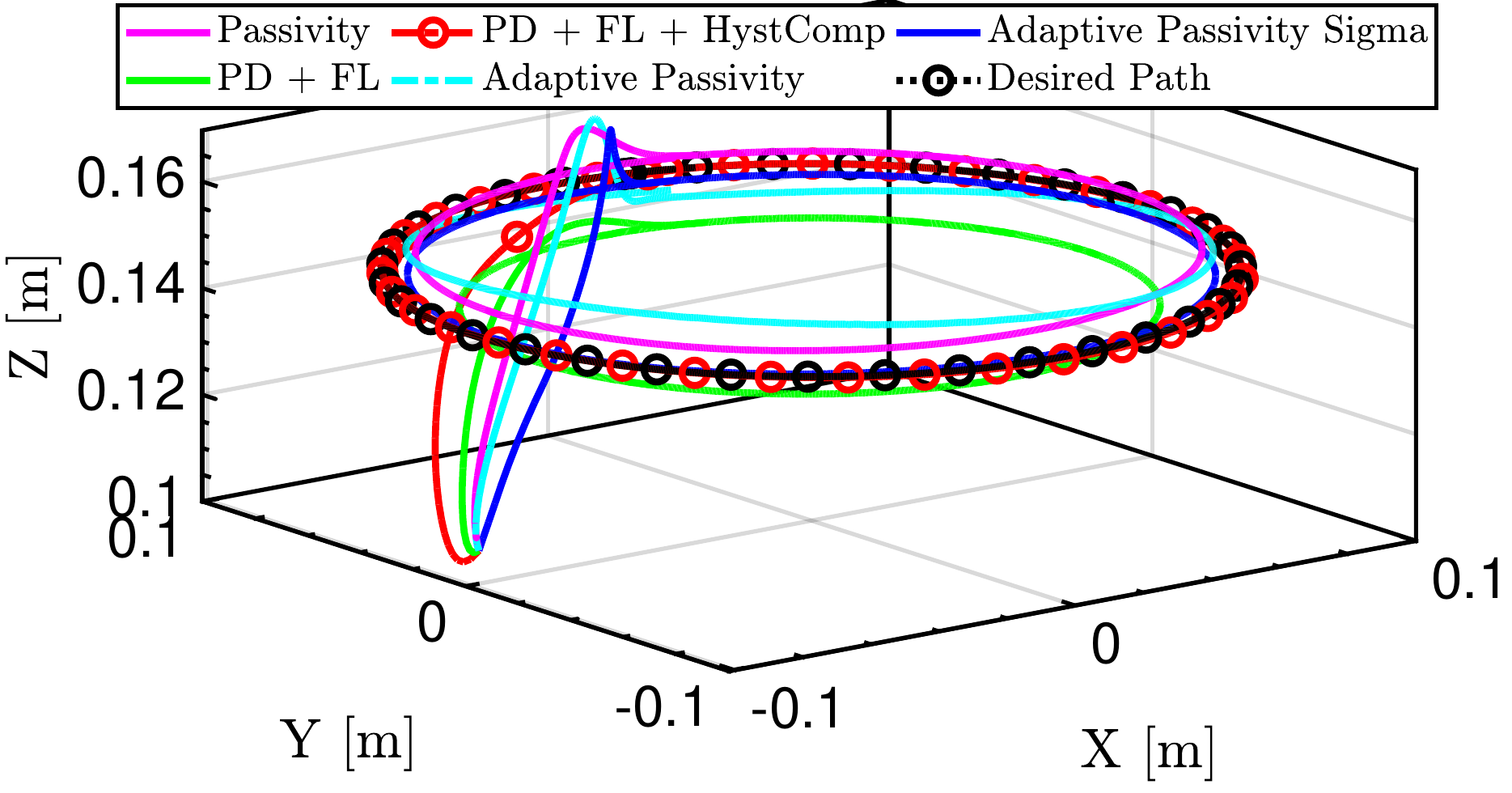}
    \caption{Trajectory tracking in practical scenarios. For a better visibility, the true position of the robot without the noise is shown in the picture}
    \label{fig:HysteresisCompFinalTrack}
\end{figure}

In this section, our proposed approach along with other mentioned control approaches are implemented in the simulation. The high-gain observer is designed based on the inaccurate model. In addition, the damping and stiffness parameters are assumed to be uncertain, the hysteresis effect is not considered in the observer, and white noise is added to the measurements of each actuator. As can be seen in Figure \ref{fig:HysteresisCompFinalTrack}, the ideal case where the model is identified exactly, the PD+FL with hysteresis compensation controller, has shown the best accuracy, which demonstrates the benefits of having an accurate model. It should be noted since the model is accurate, smaller gains are chosen for the high-gain observer. However, having an accurate model is typically not available in most soft robotic systems since the soft robot hardware is inherently complex and can be nonlinear. 

\begin{figure}[htb]
    \centering
    \includegraphics[width=0.38\textwidth]{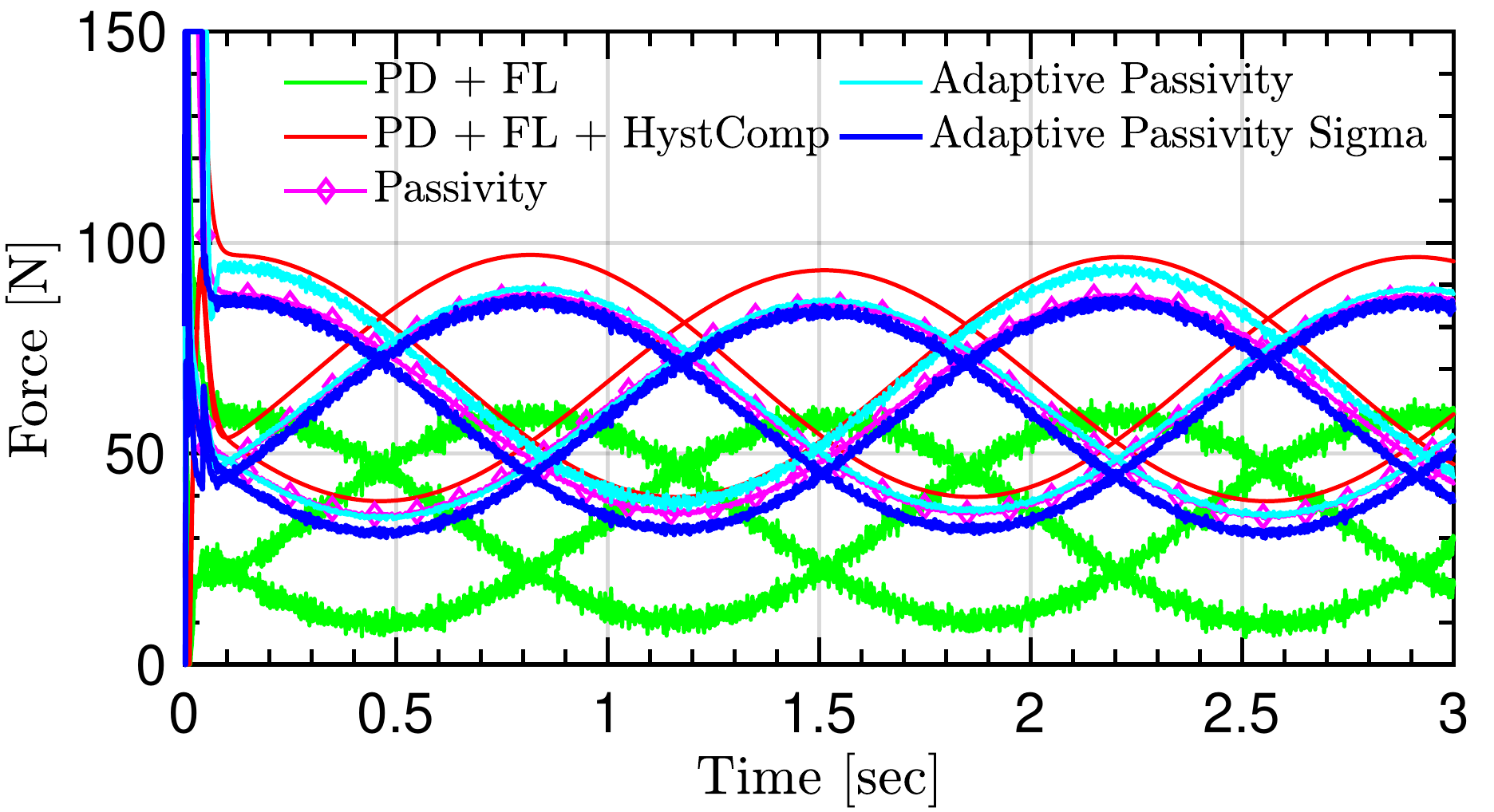}
    \caption{Control output comparison of different controllers  in practical scenario}
    \label{fig:HysteresisCompFinalControl}
\end{figure}

\begin{figure}[htb]
    \centering
    \includegraphics[width=0.38\textwidth]{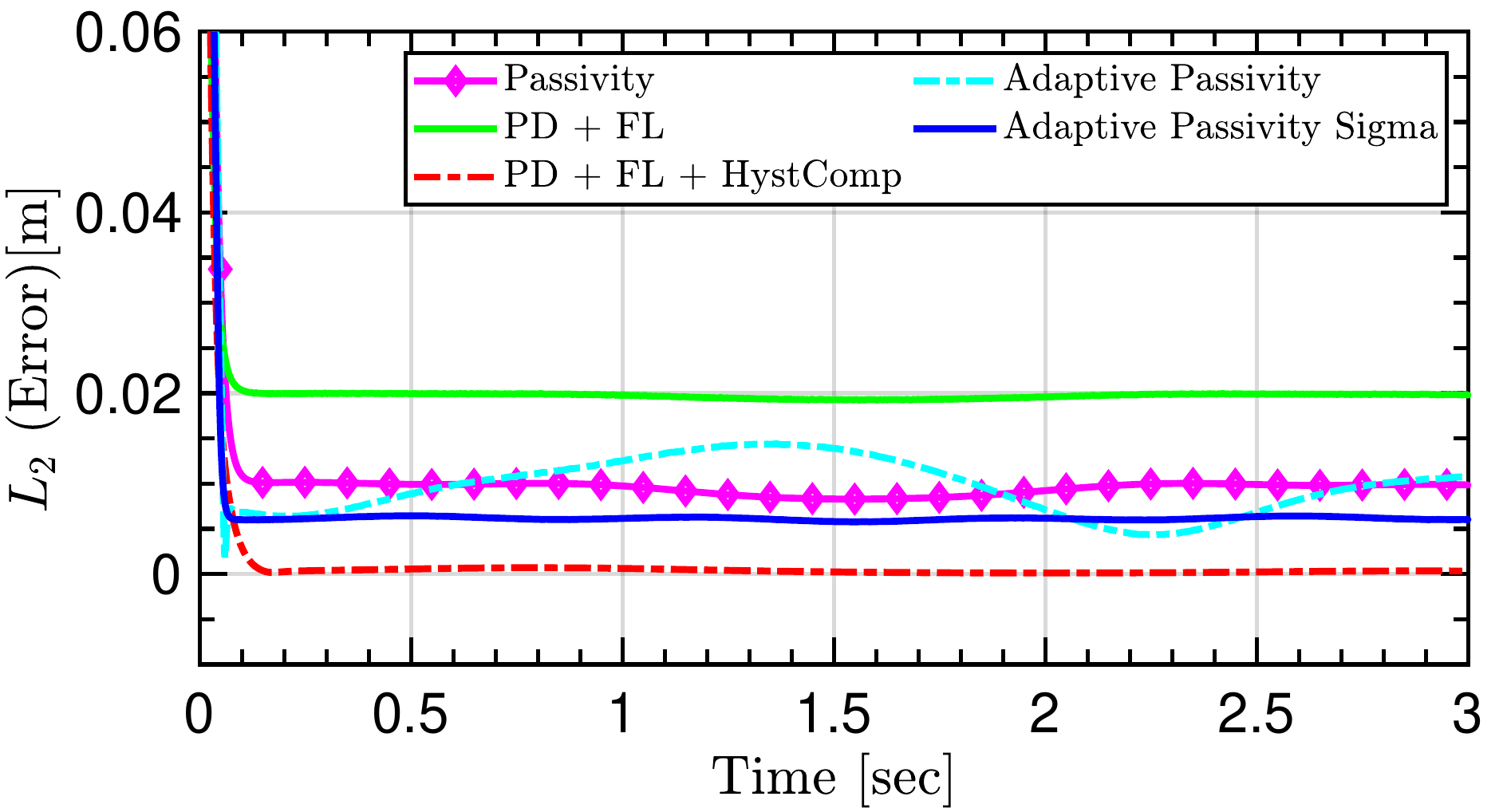}
    \caption{The $L_2$ norm comparison of different controllers in the Cartesian space in practical scenarios where hysteresis is present. The measurement of velocity is not available and white noise is present in sensor measurements.}
    \label{fig:HysteresisCompFinalL2}
\end{figure}

As can be seen in Figure \ref{fig:HysteresisCompFinalControl}, the PD+FL controller in the real-world scenario generated a control signal that includes a high chattering phenomenon since PD+FL controller doesn't have any filtered effect to reduce the sensor measurements in the control signal. For instance, the passivity based controllers have shown smoother control signals for their  error filtering structure. Nevertheless in the adaptive passivity control case, since the controller is relying on the measurements rather than the model, the effect of sensor noise would be increased. Therefore, a smaller adaptation gain is chosen for this simulation $\Gamma(i) = 1000$. Based on Figure \ref{fig:HysteresisCompFinalL2} the smaller adaptation gain version didn't improve the accuracy in comparison with the simple passivity controller. This can be associated with smaller adaptation gain and parameter drift.

The adaptive-passivity control with sigma modification could solve the above-mentioned challenges and it not only could help us to choose a bigger adaptation gain $\Gamma(i)=10^{6}$ along with a sigma modification constant $\sigma(i) = 10^{3}$, but also eliminates the effect caused by the sensor noise in the generated control signal. As can be seen in Figure \ref{fig:HysteresisCompFinalTrack}, and Figure \ref{fig:HysteresisCompFinalL2} (time-varying $L_2$ norm error between the target trajectory and robot tip trajectory) the accuracy of tracking has been increased. The chattering phenomenon is smaller than the PD+FL controller, as depicted in Figure \ref{fig:HysteresisCompFinalControl}. It should be noted that by tuning the value of adaptation and sigma-modification terms, the balance between accuracy and smooth control output can be achieved.

\section{Conclusion}
\label{sec:conclusion}
In this paper, we addressed the control problem of the one-segment soft robotic arm. The importance of implementing an adaptive controller is shown based on the simulation results under different operation scenarios. The performance of conventional control approaches is explained and the advantage and disadvantages of each approach are discussed in detail. It has been shown that the adaptive-passivity based controller outperforms other approaches and tracks the desired path with the best accuracy. Then, the practical scenario, which takes into account the hysteresis effect, parametric uncertainty, unavailability of the velocity measurements, and sensor noise, is considered in the simulation study. The adaptive-passivity based controller is modified to compensate those effects with a robustification (sigma-modification) term in the adaptation rule, along with a high-gain observer that can not only compensate the noise of the measurements but also estimate the unavailable states of the system. In our future work, we will implement the controller in practical robot hardware and multi-section soft robotic arm setup to perform more complicated tasks such as grasping  \cite{arachchige2020novel} or locomotion \cite{arachchige2020modeling}.


%

\appendices




\ifCLASSOPTIONcaptionsoff
  \newpage
\fi



\bibliographystyle{IEEEtran}
\bibliography{DynamicControlSoftArm}

\end{document}